\documentclass[review]{elsarticle}

\usepackage{lineno,hyperref}
\usepackage{multirow}
\usepackage{fancyhdr} 

\modulolinenumbers[5]

\makeatletter
\def\ps@pprintTitle{%
 \let\@oddhead\@empty
 \let\@evenhead\@empty
 \def\@oddfoot{}%
 \let\@evenfoot\@oddfoot}
\makeatother

\fancypagestyle{firststyle}
{
   \fancyhf{}
   \fancyfoot[C]{\footnotesize Journal of Information Processing \& Management. doi:10.1016/j.ipm.2015.05.003 \copyright\ 2016. This manuscript version is made available under the CC-BY-NC-ND 4.0 license http://creativecommons.org/licenses/by-nc-nd/4.0/}
}

\journal{Information Processing \& Management. doi:10.1016/j.ipm.2015.05.003 - This manuscript version is made available under the CC-BY-NC-ND 4.0 license http://creativecommons.org/licenses/by-nc-nd/4.0/}

\biboptions{authoryear}

\newenvironment{myquote}{\list{}{\leftmargin=0.05in\rightmargin=0.05in}\item[]}{\endlist}

\begin{document}

\begin{frontmatter}

\title{Why Do Urban Legends Go Viral?}

\author{Marco Guerini$^\dagger$ and  Carlo Strapparava$^{\dagger}$}
\address{$^\dagger$FBK-irst, Trento, Italy}

\begin{abstract}
Urban legends are a genre of modern folklore, consisting of stories about rare and exceptional events, just plausible enough to be believed, which tend to propagate inexorably  across communities.
In our view, while urban legends represent a form of ``sticky'' deceptive text, they are marked by a tension between the credible and incredible. They should be credible like a news article and incredible like a fairy tale to go viral.  
In particular we will focus on the idea that urban legends should mimic the details of news (\emph{who}, \emph{where}, \emph{when}) to be credible, while they should be emotional and readable like a fairy tale to be catchy and memorable. Using NLP tools we will provide a quantitative analysis of these prototypical characteristics. We also lay out some machine learning experiments showing that it is possible to recognize an urban legend using just these simple features.
\end{abstract}


\end{frontmatter}

\thispagestyle{firststyle}


\section{Introduction}

Urban legends are a genre of modern folklore consisting of stories told as true -- and plausible enough to be believed -- about some rare and exceptional events that supposedly happened to a real person or in a real place.

Whether urban legends are produced by individual authors or emerge spontaneously, they typically spread ``virally" across communities 
and tend to change over time with repetition and embellishment,
like memes \citep{dawkins2006selfish}. 
For example the sewer alligator, that originally ``appeared'' in New York City \citep{coleman1979alligators}, also appeared in different cities to suit regional variations.  
Though it is considered synonymous of ``false belief," the term urban legend refers to a subtler and more complex phenomenon. The crucial factor is that the story is told as true in the absence of verification. Folklorists are generally more interested in the social context and meaning of urban legends than their truth value. From an NLP point of view, instead, it is interesting to computationally explore those linguistic characteristics that make them appealing and bring people to circulate them. 
With the advent of the Internet, urban legends gained new lifeblood, as they began to be circulated by e-mail. 

In \citep{heath2007made}, the authors discuss the idea of ``stickiness" popularized by the book ``The Tipping Point'' \citep{gladwell2000tipping}, seeking to explain what makes an idea or concept memorable or interesting. They also focus on urban legends and claim that, by following the acronym ``SUCCES" (each letter referring to a characteristic that makes an idea ``sticky"), it is possible to describe their \emph{prototypical} structure:

\begin{itemize}
\item Simple -- find the core of any idea
\item Unexpected -- grab people's attention by surprising them
\item Concrete -- make sure an idea can be grasped and remembered later
\item Credible -- give an idea believability
\item Emotional -- help people see the importance of an idea
\item Stories -- empower people to use an idea through narrative
\end{itemize}

Such features are allegedly placed at the core of persuasive and viral language;
 urban legends constitute an ideal framework with which to computationally verify these assertions. Table \ref{tab_UL} displays a few examples of urban legends claims. 

In particular we will investigate some of the prototypical characteristics that can be found in urban legends as compared to similar literary genres.  
In our view, urban legends are viral since they are stressed by a tension between credible and incredible: credible like a \emph{news} and incredible like a \emph{fairy tale}.  
We will focus on the idea that urban legends should mimic the details of news (\emph{who}, \emph{where}, \emph{when}) to be credible, and they should be \emph{emotional}  and \emph{readable} like the story of a fairy tale to be catchy and memorable. We will verify these psychological hypotheses -- appeared in the literature -- using NLP tools, to drive a \emph{quantitative} analysis of these \emph{qualitative} theories. 
For example, the idea that urban legends derive much of their credibility from details concerning the location where the situation took place, is presented in \citep{brunvand1981vanishing}. Anecdotically, the  television series ``1000 Ways to Die" -- that recreates unusual supposed deaths and debunked urban legends in a way similar to the Darwin Awards\footnote{The Darwin Awards are an ironical honor, granted to individuals who have contributed to human evolution by ``self-selecting themselves out of the gene pool" via incredibly foolish actions; Darwin Awards explicitly try to disallow urban legends from the awards. See \texttt{darwinawards.com}} -- introducing each story with the location and date of each supposed incident, to render it more credible.

\begin{table}[htb]
\centering
{
{\small \begin{tabular}{p{8.5cm}}
\hline
A tooth left in a glass of Coca-Cola will dissolve overnight.\\
\hline
A stranger who stopped to change a tire on a disabled limo was rewarded for his efforts when the vehicle's passenger, Donald Trump, paid off his mortgage. \\
\hline
Walt Disney arranged to have himself frozen in a cryonic chamber full of liquid nitrogen upon his death, and he now awaits the day when medical technology makes his re-animation possible.\\
\hline
Drugged travelers awaken in ice-filled bathtubs only to discover one of their kidneys has been harvested by organ thieves.\\
\hline
Facebook users can receive a \$5,000 cash reward from Bill Gates for clicking a share link.\\
\hline
 \end{tabular}}
}
\caption{Examples of Urban Legend Claims}\label{tab_UL}
\end{table}

In the tension between credible and incredible, details should be neither too specific, like in the news, nor too few, as in fairy tales: effective urban legends should be credible but not verifiable. Similarly, emotions should be enough to make it sticky/catchy but not too much to render it not-credible. Finally urban legends should be easy to read, similar to fairy tales, to render them more memorable.  As an example consider the following excerpt, taken from the ``Kidney Theft'' urban legend, as reported by \texttt{snopes.com}:

\begin{myquote}
{\footnotesize  Dear Friends:\\ 
I wish to warn you about a new crime ring that is targeting business travelers. This ring is well organized [\ldots] and is currently in most major cities and recently very active in New Orleans. The crime begins when a business traveler goes to a lounge for a drink [\ldots] A person in the bar walks up as they sit alone and offers to buy them a drink. The last thing the traveler remembers until they wake up in a hotel room bath tub, their body submerged to their neck in ice, is sipping that drink. There is a note taped to the wall instructing them not to move and to call 911. [\ldots] The business traveler is instructed by the 911 operator to very slowly and carefully reach behind them and feel if there is a tube protruding from their lower back. The business traveler finds the tube and answers, ``Yes." The 911 operator tells them to remain still, having already sent paramedics to help. The operator knows that both of the business traveler's kidneys have been harvested. This is not a scam, it is real. It is documented and confirmable. If you travel, please be careful.

Regard\\
Jerry Mayfield}
\end{myquote}

There is no very strong emotional wording in this example, it is the situation itself that is scary; on the contrary the email contains locations, the signature of a presumed Jerry Mayfield, and -- noticeably -- credibility is also explicitly addressed in the text with the adjectives ``real'', ``documented'' and ``confirmable''.

\medskip

In the following sections we first review relevant work that addresses the problem of deceptive language and behavior both in online and offline scenarios, followed by an overview of work that addresses the virality of online content. Then we describe the data collected for our experiments and the features extracted to model the aforementioned prototypical characteristics of urban legends. We use these features in both descriptive statistics and generalization tasks and we report the best performing features. Finally we discuss future research on further prototypical characteristics of urban legends.

\section{Related Work}

The topic of deceptive and/or false messages is a burning topic within the NLP community. A seminal work on the linguistic recognition of lies can be found in \citep{rada:carlo:ACL-09}. Still, defense from subtle persuasive language in broadcast messages, including social networks, is needed in many applied scenarios.  Viral messages have become a very important factor for persuasion and are currently almost entirely out of control. So, protection from fraudulent communication is needed, especially in competitive commercial situations. Two main approaches are currently under investigation in the literature: 

1) Recognizing the linguistic characteristics of deceptive content in the social web: for example preventing deceptive consumer reviews \citep{Ott:2011:FDO:2002472.2002512} on sites like Trip Advisor is fundamental both for consumers seeking genuine reviews, and for the reputation of the site itself. Deceptive consumer reviews are fictitious opinions that have been deliberately written to sound authentic.  Another example concerns online advertising \citep{sculley2011detecting}: detecting fraudulent ads is in the interest of users, of service providers (e.g.\  Google AdWords system), and other advertisers. An interesting phenomenon at the crossroad of viral phenomena and deceptive customer reviews, where ironic reviews (such as the case of the mountain three wolf moon) create phenomena of social contagion, is discussed in \cite{reyes2012making}. 

2) Recognizing on-line behavioral patterns of deceptive users: For example recognizing groups of propagandists or fake accounts that are used to push the virality of content \citep{lumezanu2012bias}. Four main patterns are recognized: (i) sending high volumes of tweets over short periods of time, (ii) retweeting while publishing little original content, (iii) quickly retweeting, and (iv) colluding with other, seemingly unrelated, users to send duplicate or near-duplicate messages on the same topic simultaneously. Another example is \citep{feng2012distributional} where the authors hypothesize that there is a set of representative distributions of review rating scores. Deceptive business entities that hire people to write fake reviews can then be recognized since they will necessarily distort distribution of review scores, leaving ``distributional footprints" behind. 

We want to consider a third point, which is linked to the previous two but different at the same time: deceptive content that spreads quickly but without an explicit strategy of making them spread, which is the case with urban legends. 

Finally, the spreading dynamics of an urban legend on one hand closely resembles those of memes that undergo many variations while spreading \citep{simmons2011memes}; on the other hand their characteristics resemble those of viral content. 
Several researchers have studied information flow, community building and similar processes using Social Networking sites 
as a reference \citep{contagionDigg,caseStudyOnComments,credibility}. However, the great 
majority concentrate on network-related features without taking into account the actual content spreading within the network \citep{voting}.
A hybrid approach focusing on both product characteristics and network related features is presented in \citep{aral2011creating}:
in particular, the authors study the effect of passive-broadcast and active-personalized notifications embedded in an application 
to foster word of mouth.

Recently, the correlation between content characteristics and virality has begun to be investigated, especially 
with regard to textual content; in \citep{opinion}, for example,
features derived from sentiment analysis of comments are used to predict stories' popularity. 
The work in \citep{virality} uses \textit{New York Times} articles to examine the relationship between 
emotions evoked by the content and virality, using semi-automated sentiment analysis to quantify the affectivity 
and emotionality of each article. Results suggest a strong relationship between affect and virality, 
where virality corresponds to the number of times the article was email forwarded.  

The relevant work in \citep{danescu2012you} measures a different form of content 
spreading by analyzing which features of a movie quote make it ``memorable" online. 
Another approach to content virality, somehow complementary to the previous one, is presented in \citep{simmons2011memes}, 
and takes the perspective of understanding which modification dynamics make a meme spread from one person to another 
(while movie quotes spread remaining exactly the same).  
More recently, some works tried to investigate how different textual contents give rise to different reactions 
in the audience: the work presented in \citep{marco:carlo:gozde:ICWSM-11}  correlates several viral 
phenomena with the wording of a post,  while \citep{guerini2012linguistic} shows that  specific content features 
variations (like the readability level of an abstract) differentiate among virality level of downloads, 
bookmarking, and citations. 

\section{Datasets}

To explore the characteristics of urban legends and understand the effectiveness of our ideas we collected a specific dataset. It is composed of roughly 8000 textual examples: 2518 Urban Legends (UL), 1860 Fairy Tales (FT) and 3575 Google News articles (GN). The description of how the datasets have been created follows.

\begin{itemize}
\item \textbf{Urban Legends} have been harvested from the website \texttt{snopes.com}. While almost 5 thousand urban legends were collected and discussed on the website, we considered only those that were reported along with a textual example, usually e-mail circulated on the Internet\footnote{In our dataset, roughly 60\% of the cases are emails, 40\% are examples collected from other sources (e.g.\ websites, local newspapers, forums).}. We then kept only those textual examples  when at least thirty tokens long. 

\item \textbf{News Articles} have been selected from a corpus of about 400.000 Google News articles, from the years 2009-2012. We collected
those with the highest similarity among the titles of the Urban Legends, to grant that textual content is comparable. The similarity scores were computed in a Latent Semantic space, built from the
British National Corpus using 400 dimensions. The typical categories of GN articles are science, health, entertainment, economy and sports, news from world.

\item \textbf{Fairy Tales} We exploit a corpus of fairy tales collected and preprocessed by \citep{lobo2010fairy} that were downloaded from Project Gutenberg \citep{hart2000project}.
Since the corpus ranges from very short tales (the shortest is 75 words) to quite long ones (the longest is 15,000 words)
we split the longest tales to get a total of 1860 documents. The mean length of the resulting documents is about 400 words.
\end{itemize}

\section{Feature Extraction}

After collecting the datasets we extracted four different groups of features, relevant to the prototypical characteristics we want to analyze.

\textbf{Named Entities}, (\emph{NE}). To annotate named entities we used the TextPro toolkit \citep{pianta2008textpro}, and in particular its Named Entities recognition module. 
The output of the tool is in the IOB2 format and includes the tags Person (\emph{PER}), Organization (\emph{ORG}), Location (\emph{LOC}) and Miscellaneous (\emph{MISC}).

\textbf{Temporal Expressions}, (\emph{TIMEX}). To annotate temporal expressions we used the toolkit  TTK \citep{verhagen2008temporal}. The output of the tool is in TimeML annotation language format  \citep{pustejovsky2003timeml}. In particular time expressions are flagged with TIMEX3 tags (tern.mitre.org). 
The tags considered are \emph{DATE}, \emph{DURATION} and \emph{TIME}.

To compute the importance of the aforementioned features, and to explore the characteristics of urban legend texts, we used the method proposed in \citep{rada:carlo:ACL-09}. We calculate a score associated with a given set of entities (features), as a measure of saliency for the given word class inside the text, called \emph{coverage}.

More formally, given a set of feature instances present in a text, C$=$\{$W_1$, $W_2$, \ldots, $W_N$\}, we define the feature coverage in that text (or corpus)
$A$ as the percentage of words from $A$ belonging to the feature set $C$:

{\small
\begin{equation}
Coverage_{A}(C) =  \frac{\sum_{W_{i} \in C} Frequency_A(W_i)}{Words_{A}}
\end{equation}
}

where $Frequency_A(W_i)$ represents the total number of feature occurrences $W_i$ inside the text
A, and $Words_{A}$ represents the total size (in words) of the text.
Note that we computed features' coverage regardless of their actual length: ``New York City" or ``Paris'' both count as one \emph{LOC} even if the former is composed of three tokens while the latter only of one. Note also that this approach normalizes according to text length, avoiding biases due to different corpus characteristics. \\

\textbf{Sentiment} (\emph{SENT}). Since the three corpora have different characteristics, rather than computing word polarity using specialized bag-of-words approaches, we resort to words' \emph{prior polarity} - i.e.\  if a word out of context evokes something positive or something negative. This technique, even if less precise, guarantee that the same score is given to the same word in different contexts, and that none of the corpora is either overestimated or underestimated. To this end, we follow the methodology proposed in \citep{gatti2012assessing}, using SentiWordNet 3.0 \citep{Esuli06}, that assigns prior polarities to words starting from their posterior polarities. In particular we choose the 
best performing approach. This formula uses a weighted mean, i.e.\  each sense weight is chosen according to a harmonic series. The rationale behind this choice is based on the assumption that more frequent senses should bear more ``affective weight'' than very rare senses when computing the prior polarity of a word. In particular, for each word we returned its positive (\emph{POS}) and negative (\emph{NEG}) prior polarity score:  

\begin{equation}
 \qquad POS =\frac{\sum_ {i=1}^{n}( \frac{1}{i} \times posScore_ {i})}
                                      {\sum_ {i=1}^{n}( \frac{1}{i})}
	\label{eq:wnweight2}
\end{equation}

where $posScore_ {i}$ represents the modulus of the positive polarity of the ith sense of that word. The \emph{NEG} score is computed following the same procedure.

\textbf{Emotions} (\emph{EMO}). 
To sense emotions from text we used the methodology described in \citep{carlo:rada:ACM-SAC-08}. 
The idea underlying the method
is the distinction between \emph{direct} and \emph{indirect} affective words.
For direct affective words, we refer to the WordNet Affect
\citep{strappaLREC04} lexicon, an extension of the WordNet
database which employs six basic emotion labels
(anger, disgust, fear, joy, sadness, surprise) to annotate WordNet
synsets. LSA is then used to learn, in an unsupervised setting, a
vector space from the British National Corpus. 
In the LSA space, each emotion label can be represented in various way. In particular, we employ
the `LSA Emotion Synset' setting, in which the synsets of direct
emotion words are considered. The affective load is computed in terms of its lexical similarity with
respect to one of the six emotion labels. The overall affective
load of a text is then calculated as the average of its
similarity with each emotion label.

Emotions and Sentiment features are grouped under the label \textbf{Affect} (\emph{AFF}).
\medskip

\textbf{Readability} (\emph{READ}). We further analyzed the texts in the three datasets according to readability indices, to understand whether there is a difference in the language difficulty among them. Basically, the task of readability assessment consists of quantifying how difficult a text is for a reader. This kind of assessment has been widely used for several purposes, such as evaluating the reading level of children and impaired persons and improving Web content accessibility, see for example what reported in \citep{tonelli2012making}. 

We use three indices to compute the difficulty of a text: the
Gunning Fog \citep{gunning-52}, Flesch \citep{flesch-46} and Kincaid \citep{kincaidfishburrogers:75} indices. These metrics combine factors such as word and sentence length that are easy to compute and approximate the linguistic elements that have an impact on readability. In the following formulae, $Sent_A$ represents the number of sentences in text $A$, $Cpx_A$ the number of complex words (those with three or more syllables), and $Syll_A$ the total number of syllables. 

The \emph{Fog index }is a rough measure of how many years of
schooling it would take someone to understand the content; higher
scores indicate material that is harder to read. 
Texts requiring near-universal understanding have an index less than 8. Academic papers usually have a score between 15 and 20. 
The score, for a given text $A$, is calculated according to the formula:

{\small
\begin{equation}
       Fog_{A} = 0.4 \Big( \frac{Words_A}{Sent_A} + 100 \frac{Cpx_A}{Words_A} \Big)
	\label{eq:fog}
\end{equation}
}

The \emph{Flesch Index }rates  texts on a 100-point scale. Higher
scores indicate material that is easier to read while lower numbers
mark passages that are more difficult to read. Scores can be
interpreted as: 90-100 for content easily understood by an average
11-year-old student, while 0-30 for content best understood by university graduates. The score is calculated with the following formula:

{\small
\begin{equation}
       Flesch_{A} = 206.835 - 1.015 \frac{Words_A}{Sent_A} -84.6 \frac{Syll_A}{Words_A} 
	\label{eq:flesch}
\end{equation}
}

The \emph{Kincaid Index}  or ``Flesch--Kincaid Grade Level Formula" translates the 0-100 score of the Flesch Index to a U.S. grade level. It can be interpreted as the number of years of education required to understand this text, similar to the Gunning Fog index. The grade level is calculated with the following formula:

{\small
\begin{equation}
       Kincaid_{A} = 0.39 \frac{Words_A}{Sent_A} + 11.8 \frac{Syll_A}{Words_A} - 15.59
	\label{eq:kincaid}
\end{equation}
}

\section{Descriptive Statistics}

As can be seen from Tables \ref{tab_NER} (Named Entities) and \ref{tab_TIMEX} (Temporal Expressions), urban legends  place half-way between fairy tales and news, as we expected. 
While fairy tales represent out-of-time, out-of-place and always-true stories (``\emph{a long time ago in a faraway land}"), news represent circumstantial description of events. This is reflected by the overall use of named entities (respectively almost three and four times more in UL and GN) and of temporal expressions (respectively almost two and three times more). Interestingly the use of person names is the only case where FT reduce the lead of UL and GN, and can be explained by the fact that characters in FT are usually addressed with proper names (e.g. ``\emph{Hansel} and \emph{Gretel}''). 



\begin{table*}[htb]
\centering
{\small
 \begin{tabular}{l|rr|rr|rr|rr|rr}
\hline
& \multicolumn{2}{c}{PER} & \multicolumn{2}{c}{LOC} & \multicolumn{2}{c}{ORG} & \multicolumn{2}{c}{MISC} & \multicolumn{2}{c}{ALL}\\ 
& \multicolumn{1}{c}{$\mu$} & \multicolumn{1}{c}{$\sigma$} & \multicolumn{1}{c}{$\mu$} & \multicolumn{1}{c}{$\sigma$} & \multicolumn{1}{c}{$\mu$} & \multicolumn{1}{c}{$\sigma$} & \multicolumn{1}{c}{$\mu$} & \multicolumn{1}{c}{$\sigma$} & \multicolumn{1}{c}{$\mu$} & \multicolumn{1}{c}{$\sigma$} \\
\hline
FT & 0.86\% & 1.03\% & 0.31\% & 0.44\% & 0.27\% & 0.45\% & 0.15\% & 0.30\% & 1.58\% & 1.42\% \\
UL & 1.29\% & 1.49\% & 1.12\% & 1.40\% & 1.17\% & 1.56\% & 0.57\% & 0.90\% & 4.15\% & 3.18\% \\
GN & 1.65\% & 1.19\% & 2.02\% & 1.59\% & 1.63\% & 1.28\% & 0.93\% & 0.95\% & 6.22\% & 2.91\% \\
\hline
Average & 1.26\% & 1.24\% & 1.15\% & 1.14\% & 1.02\% & 1.10\% & 0.55\% & 0.72\% & 3.98\% & 2.50\% \\
\hline
 \end{tabular}
}\caption{Coverage of Named Entities}\label{tab_NER}
\end{table*}



\begin{table*}[htb]
\centering
{\small
 \begin{tabular}{l|rr|rr|rr|rr}
\hline
& \multicolumn{2}{c}{DATE} & \multicolumn{2}{c}{DURATION} & \multicolumn{2}{c}{TIME} & \multicolumn{2}{c}{ALL}\\ 
& \multicolumn{1}{c}{$\mu$} & \multicolumn{1}{c}{$\sigma$} & \multicolumn{1}{c}{$\mu$} & \multicolumn{1}{c}{$\sigma$} & \multicolumn{1}{c}{$\mu$} & \multicolumn{1}{c}{$\sigma$} & \multicolumn{1}{c}{$\mu$} & \multicolumn{1}{c}{$\sigma$}\\
\hline
FT & 0.42\% & 0.39\% & $\:\;$0.08\% & 0.16\% & *0.01\% & 0.05\% & 0.51\% & 0.43\% \\
UL & 0.70\% & 0.93\% & *0.20\% & 0.43\% & *0.02\% & 0.12\% & 0.92\% & 1.08\% \\
GN & 1.20\% & 0.85\% & *0.20\% & 0.32\% & $\:\;$0.06\% & 0.20\% & 1.46\% & 0.96\% \\
\hline
Average & 0.77\% & 0.72\% & $\:\;$0.16\% & 0.30\% & $\:\;$0.03\% & 0.12\% & 0.96\% & 0.82\% \\
\hline
 \end{tabular}
}\caption{Coverage of Temporal Expressions}\label{tab_TIMEX}
\end{table*}

In Table \ref{tab:SENT}, statistics for sentiment and emotion coverage are reported. As can be seen, in the \emph{SENT} group of features the differences are less marked and, quite surprisingly, ULs have the lowest scores. As we would expect, FTs have the highest score. Sentiment does not meet our initial expectation and seems in contrast with previous works -- see for example what reported in \citep{heath2001emotional} on UL and evoked emotions; still 
the results on sentiment as compared to emotions can be explained by 
the distinction between \emph{affective impact} and \emph{affective language}. In fact, affective impact can either derive from the wording of the text itself (usage of strong affect words), or from the depicted situation (i.e.\  emotions are evoked by describing a vivid situation with a plain language). In our experiment we tested the `wording'  using \emph{SENT} features while the `evoked emotions'  with the \emph{EMO} features. So, UL seem to use a plain and objective language, similar to GN, to gain credibility, but tend to evoke strong emotions (similar to FT) to be catchy.
Let us consider the ``Kidney Theft'' excerpt described in Section 1, as stated, there is no very strong emotional wording in this UL, it is the depicted situation that is scary \emph{per se}. 

%

%

\begin{table}[h!]
\centering
\small
\begin{tabular}{l|rr|rr|rr}
\hline
& \multicolumn{2}{c}{POS\_sent} & \multicolumn{2}{c}{NEG\_sent} & \multicolumn{2}{c}{ALL\_sent}  \\
& \multicolumn{1}{c}{$\mu$} & \multicolumn{1}{c}{$\sigma$} & \multicolumn{1}{c}{$\mu$} & \multicolumn{1}{c}{$\sigma$} & \multicolumn{1}{c}{$\mu$} & \multicolumn{1}{c}{$\sigma$} \\
\hline
FT & 2.40 & 0.69 & 2.62 & 0.69 & 5.03 & $\;\;$1.12 \\
UL & 2.13 & 0.74 & 2.20 & 0.90 & 4.33 & *1.35 \\
GN & 2.27 & 0.87 & 2.29 & 0.73 & 4.56 & *1.33 \\
\hline
$\sigma$ & 2.27 & 0.77 & 2.37 & 0.77 & 4.64 & $\;\;$1.27 \\
\hline
& \multicolumn{2}{c}{POS\_emo} & \multicolumn{2}{c}{NEG\_emo} & \multicolumn{2}{c}{ALL\_emo}\\
& \multicolumn{1}{c}{$\mu$} & \multicolumn{1}{c}{$\sigma$} & \multicolumn{1}{c}{$\mu$} & \multicolumn{1}{c}{$\sigma$} & \multicolumn{1}{c}{$\mu$} & \multicolumn{1}{c}{$\sigma$} \\
 \hline
FT & 0.63 & 0.03 & 0.58 & 0.04 & 0.60 & 0.03 \\
UL & 0.54 & 0.07 & 0.49 & 0.07 & 0.51 & 0.06 \\
GN & 0.50 & 0.06 & 0.48 & 0.06 & 0.48 & 0.06 \\
\hline
$\sigma$ & 0.56 & 0.05 & 0.52 & 0.06 & 0.53 & 0.05 \\
\hline
\end{tabular}
\vspace{-0.3cm}
\caption{Average Sentiment and Emotions}\label{tab:SENT}
\end{table}

In Table \ref{tab_READ}, statistics for readability are reported. As can be seen, ULs are readable in a way similar to fairy tales. Still, depending on the readability indices, that grasp different aspects of text difficulty,  ULs are either slightly easier than FTs or half-way between FTs and ULs similar to the cases of Tables 
\ref{tab_NER} and \ref{tab_TIMEX}.

\begin{table}[h]
\centering
{\small
\begin{tabular}{l|rr|rr|rr|rr|rr|rr}
\hline
& \multicolumn{2}{c}{Fog} & \multicolumn{2}{c}{Flesch} & \multicolumn{2}{c|}{Kincaid} & \multicolumn{2}{c}{ $\frac{Cpx_A}{Words_A}$} & \multicolumn{2}{c}{$\frac{Syll_A}{Words_A}$} & \multicolumn{2}{c}{$\frac{Words_A}{Sent_A}$}  \\

& \multicolumn{1}{c}{$\mu$} & \multicolumn{1}{c}{$\sigma$} & \multicolumn{1}{c}{$\mu$} & \multicolumn{1}{c}{$\sigma$} & \multicolumn{1}{c}{$\mu$} & \multicolumn{1}{c|}{$\sigma$} & \multicolumn{1}{c}{$\mu$} & \multicolumn{1}{c}{$\sigma$} & \multicolumn{1}{c}{$\mu$} & \multicolumn{1}{c}{$\sigma$} & \multicolumn{1}{c}{$\mu$} & \multicolumn{1}{c}{$\sigma$}\\
\hline
FT & 12.71 & $\:\;$3.68 & 67.04 & 10.92 & 10.37 & $\:\;$3.48 & $\:\:$5.77 & 2.41 & 1.34 & 0.07 & 26.01 & $\:\;$8.43 \\
UL & 12.13 & *5.00 & 58.56 & 17.31 & $\:\;$9.65 & *4.70 & 11.67 & 5.13 & 1.51 & 0.19 & 18.74 & 10.96 \\
GN & 15.69 & *4.95 & 41.10 & 16.39 & 12.90 & *4.64 & 17.26 & 4.68 & 1.67 & 0.21 & 22.12 & 10.77 \\
\hline
$\mu$ & 13.51 & $\:\;$4.54 & 55.57 & 14.87 & 10.98 & $\:\;$4.28 & 11.57 & 4.07 & 1.51 & 0.15 & 22.29 & 10.05 \\
\hline
 \end{tabular}
}\caption{Readability difficulty Scores}\label{tab_READ}
\end{table}


This behavior can be explained by the fact that ULs have a simpler syntax than FTs but a more complex lexicon. In fact, inspecting the individual elements of the formulae, as reported in the second part of Table \ref{tab_READ}, we see that while the percentage of complex words (either $\frac{Cpx_A}{Words_A}$ or $\frac{Syll_A}{Words_A}$) puts UL halfway between FT and GN, the average length of sentences ($\frac{Words_A}{Sent_A}$) is surprisingly higher for FT than GN and in turn UL. So, depending on the weight given either to complex words or to sentence length, the results in Table \ref{tab_READ} can be interpreted. 

%

All differences in the means reported in the tables are statistically significant (Student's t-test, $p<0.001$) apart from 
\emph{TIME}, between UL and FT, and \emph{DURATION}, between UL and GN, (signalled with * in Table \ref{tab_TIMEX}).

Turning to the analysis of variance, we see that FT is -- on average -- a more cohesive genre, with lower standard deviations, while GN and UL have higher and closer standard deviations.  In fact, all differences in the standard deviations reported in the tables are statistically significant  (f-test, $p<0.001$) apart between UL and GN in \emph{Fog}, \emph{Kincaid} and in \emph{ALL} sentiment (signalled with * in the respective Tables). 

\section{Classification Experiments}

The goal of our experiments is to understand to what extent it is possible to assign a text to one of the aforementioned classes using just the prototypical characteristics (features) discussed above, and whether there is a subset of features that stands out among the others in this classification task. For every feature combination we conducted a binary classification experiment with ten-fold cross validation on the dataset. We always randomly downsampled the majority class in order to make the dataset balanced, i.e.\  50\% of positive examples and 50\% of negative examples; this accounts for a random baseline of 0.5. We also normalized all features according to z-score. Experiments were carried out using SVM \citep{Vapnik95}, in particular libSVM \citep{chang2011libsvm} under its default settings. Results are reported in Table \ref{tab_Classification}; all significance tests discussed below are computed using an approximate randomization test \citep{yeh2000more}.


\textbf{Urban Legends vs News}. In the UL vs.\  GN classification task, while all the features together performed well (F1 = 0.833), improving over all other subgroups of features ($p<0.001$), no single group of features performed so well, apart from READ (F1 = 0.763, $p<0.001$). Particularly, the temporal features (\emph{TIMEX}) performed worse than \emph{AFF} and \emph{NE} ($p<0.001$). Still, all features 
improved over the baseline ($p<0.001$). 

\begin{table}[h]
\centering
{
{\small \begin{tabular}{l|rrr|rrr|rrr}
\hline
& \multicolumn{3}{c|}{UL vs. GN} & \multicolumn{3}{c|}{UL vs. FT} & \multicolumn{3}{c}{GN vs. FT} \\
\hline
Features & Prec & Rec & F1 & Prec & Rec & F1 & Prec & Rec & F1  \\
\hline

NE & 0.694 & 0.694 & 0.694 & 0.787 & 0.768 & 0.777 & 0.897 & 0.896 & 0.896\\
TIMEX & 0.677 & 0.676 & 0.676 & 0.666 & 0.666 & 0.666 & 0.775 & 0.767 & 0.766\\
AFF & 0.735 & 0.735 &  0.735 & 0.865    &  0.863 &  0.864 & 0.946 & 0.946 & 0.946\\
READ & 0.765 & 0.762 & 0.763 & 0.869 & 0.868 & 0.868 & 0.973 & 0.973 & 0.973\\
\hline
ALL & 0.834 &  0.833 & 0.833  &  0.897 & 0.897 & 0.897 & 0.978 & 0.978 & 0.978 \\
\hline

Rand & 0.500 &  0.500 & 0.500  &  0.500 & 0.500 & 0.500 & 0.500 & 0.500 & 0.500 \\
\hline
 \end{tabular}}
}\caption{Classification Results}\label{tab_Classification}
\end{table}

%

\textbf{Urban Legends vs Fairy Tales}. In the UL vs.\  FT classification task, all the features together performed better than the previous experiment  (F1 = 0.897), again improving over all the other subgroups of features alone ($p<0.001$). Interestingly, the best discriminative subgroup of features (still \emph{READ}, F1 = 0.868) in this case reduces the lead with respect to all the features together (\emph{ALL}) and improves over the others subgroups ($p<0.001$) apart from the \emph{AFF} group -- from which has no significant difference -- that in this case performs better than in the previous experiment. On the contrary, the \emph{TIMEX} group had similar performances as the previous experiment, while \emph{NE} improved its performance.  Finally, all groups of features had a statistically significant improvement over the baseline ($p<0.001$).


In Table \ref{tab_single_class} we report the performances of the various classification tasks in term of precision, recall and F1 over the single classes. Interestingly, for almost all feature combinations the classifiers had slightly higher precision than recall for UL, while the contrary holds for FT and GN. 	 

\begin{table}[!h]
\centering
{
{\small \begin{tabular}{l|r|rrr}
\hline
Features & Class & Prec & Rec & F1 \\
\hline
\multicolumn{5}{c}{UL vs GN}\\
\hline
\multirow{2}{*}{ALL} & UL & 0.846 & 0.815 & 0.830 \\
& GN & 0.822 & 0.851 & 0.836\\
\hline
\multirow{2}{*}{NE} & UL & 0.699 & 0.680 & 0.689 \\
 & GN & 0.689 & 0.707 & 0.698\\
\hline
\multirow{2}{*}{TIMEX} & UL & 0.691 & 0.637 & 0.663\\
 & GN & 0.663 & 0.716 & 0.689 \\
\hline
\multirow{2}{*}{AFF} & UL & 0.727 & 0.754 & 0.740 \\
 & GN &  0.744 & 0.716 & 0.730\\
\hline
\multirow{2}{*}{READ} & UL &  0.794 & 0.708 & 0.749\\
 & GN & 0.737 & 0.816 & 0.774\\
\hline
\multicolumn{5}{c}{UL vs FT}\\
\hline
\multirow{2}{*}{ALL}  & UL & 0.910 & 0.880 & 0.895 \\
 & FT & 0.884 & 0.913 & 0.898 \\
\hline
\multirow{2}{*}{NE} & UL &  0.859 & 0.641 & 0.734 \\
 & FT &  0.714 & 0.895 & 0.794\\
\hline
\multirow{2}{*}{TIMEX} & UL  & 0.666 & 0.663 & 0.665 \\
 & FT  & 0.665 & 0.668 & 0.666\\
\hline
\multirow{2}{*}{AFF} & UL & 0.885 & 0.835 & 0.860\\
 & FT & 0.844 & 0.891 & 0.867 \\
 \hline
\multirow{2}{*}{READ} & UL &  0.883 & 0.849 & 0.865\\
 & FT & 0.854 & 0.887 & 0.870\\
\hline
 \end{tabular}}
}
\caption{Classification Results on Single Classes}\label{tab_single_class}
\end{table}

\textbf{News vs Fairy Tales}. Finally, we wanted to check whether UL being ``half-way" between GN and FT can be observed in our classification experiments as well. If this hypothesis is correct, by  classifying GN vs.\  FT we would expect to find higher performance than previous experiments. Results show that this is in fact the case. 
All features together performed better than all previous experiment and incredibly well (F1= 0.978),  again improving over all the other subgroups of features alone ($p<0.001$) apart from READ that performs equally well (F1=0.973, no statistically significant difference). 
Notably, all other groups of features improves over the UL vs.\  GN and the UL vs.\  FT  tasks. 
Finally, all groups of features had a statistically significant improvement over the random baseline ($p<0.001$).

\textbf{Three Class Classification}. Finally we also tested feature predictivity on a three class classification task (UL vs GN vs FT). Since in this case we did not performed downsampling, we use the ZeroR classifier as a baseline. For the sake of  interpretability of results, along with precision, recall and F1 we also provide the Matthews Correlation Coefficient (MCC) which is useful for unbalanced datasets, as presented in \citep{gorodkin2004comparing}  for the multiclass case. MCC returns a value between -1 and +1, where +1 represents a perfect prediction, 0 no better than random and -1 indicates total disagreement.  Results are consistent with previous experiments. In Table \ref{tab_UL-FT-GN}, all feature configurations show an improvement over the baseline ($p<0.001$) but the temporal features (\emph{TIMEX}) have far lower discriminative power as compared to others groups of features (MCC=0.339). 


\section{Discussion}

While between UL and GN the discrimination is given by a skillful mixture of all the prototypical features together, where none has a clear predominance over the others, 
between UL and FT, readability (\emph{READ}) and affect (\emph{AFF}) play a major role.  From the summary in Table \ref{tab_sum} we see that while ALL features together have the highest averaged F1, \emph{READ} is the best performing subset of features in all experiments, followed by \emph{AFF}, \emph{NER} and \emph{TIMEX} that perform reasonably well. n general, these experiments proved the goodness of our features in discriminating UL against FT and GN in a machine learning framework, confirming the results emerged from the quantitative analysis part. In particular, as expected, these features gave the best results in the GN vs FT experiments, showing that these two genres represent the extremes of a continuum where ULs are placed. 



\noindent
\begin{table*}[!htb]    
    \begin{minipage}{.57\linewidth} 
    \centering  {\small   
	\begin{tabular}{l|rrrr}
	\hline
	Features & Prec & Rec & F1 & MCC \\
	\hline
	NE & 0.630 & 0.650 & 0.640 & 0.449\\
	TIMEX & 0.570 & 0.577 & 0.573 & 0.339\\
	AFF & 0.707 & 0.717 &  0.712 & 0.552 \\
	READ & 0.746 & 0.754 & 0.750 & 0.611\\
	ALL &  0.820 & 0.822 & 0.821 & 0.721\\
	ZeroR & 0.202 & 0.450 & 0.279 & 0\\
	\hline
	\end{tabular}}\caption{Results for UL vs FT vs GN }\label{tab_UL-FT-GN}
    \end{minipage}%
    \begin{minipage}{.5\linewidth}
    \centering {\small
	\begin{tabular}{l|rr}
	\hline	
	Features & F1$\mu$ & F1$\sigma$ \\
	\hline
	ALL & 0.868 & 0.070 \\
	SENT & 0.848 & 0.106 \\
	READ & 0.819 & 0.100 \\
	NE & 0.740 & 0.100 \\
	TIMEX & 0.675 & 0.069 \\
	\hline
	\multicolumn{3}{c}{}\
	\end{tabular}}\caption{Overall Feature performances}\label{tab_sum}
   \end{minipage}
\end{table*}

\section{Ecological Experiments in the News Domain}

As a final validation of our feature importance we also set up experiments where we controlled for the medium where the message is delivered, specifically the online news domain. Since Newspapers exist for all kinds of stories and with all sorts of reputations for reliability we focused on two specific websites. One is the \emph{Weekly World News} (WWN), a news website with very low reliability where many of the stories have the qualities of urban legends (the WWN was famous for stories about Bigfoot and UFOs, etc.). The other website is \emph{The New York Times} (NYT), known for its high reliability and fact-checking procedures. 

We scraped the WWN for a total of 225 stories, and  then randomly selected an equal amount of stories from the NYT. For both datasets we extracted the same set of features discussed in the previous sections. For every feature combination we conducted a binary classification experiment with ten-fold cross validation on the dataset. Since the dataset is balanced, this accounts for a random baseline of 0.5. We also normalized all features according to z-score. Results are reported in Table \ref{tab_Classification_news}.

\begin{table}[h]
\centering
{
{\small \begin{tabular}{l|rrr}
\hline
Features & Prec & Rec & F1  \\
\hline

NE & 0.770 & 0.769 & 0.769 \\
TIMEX & 0.751 & 0.749 & 0.750\\
AFF & 0.693 & 0.680 & 0.686\\
READ & 0.774 & 0.756 & 0.765\\
\hline
ALL  & 0.865 & 0.863 & 0.864 \\
\hline
Rand & 0.500 & 0.500 & 0.500 \\
\hline
 \end{tabular}}
}\caption{Classification Results, WWN vs. NYT}\label{tab_Classification_news}
\end{table}

Also in this case our features are able to discriminate between reliable and non-reliable stories (namely those coming from NYT and WWN). In particular,  all the features together performed very well (F1 = 0.864), improving over all other subgroups of features (p $<$ 0.001), and NE, TIMEX, READ performed equally well improving over AFF that was the least effective (p $<$ 0.001). Still, AFF improves over the random baseline (p $<$ 0.001). 

With this last experiment we were able to show that stories from different newspapers of differing reliability might be classified correctly using the features learned for discriminating regular news from urban legends. So, also in more applicative and ecological scenarios,  where stories come from the same medium (online news) these features are useful in discriminating stories on the basis of their UL-ness or GN-ness.

\section{Conclusions}

In this paper we have presented a study on urban legends, a genre of modern folklore consisting of stories about some rare and exceptional events plausible enough to be believed.  We argued that urban legends represent a form of ``sticky'' deceptive text, marked by a tension between the credible and incredible. To be credible they should resemble a news article while being incredible like a fairy tale.  In particular we focused on the idea that ULs should mimic the details of news (\emph{who}, \emph{where}, \emph{when}) to be credible, while being \emph{emotional} and \emph{readable} like a fairy tale to be catchy and memorable. Using NLP tools we presented a quantitative analysis of these simple yet effective features and provided some machine learning experiments showing that it is possible to recognize an urban legend using just these prototypical characteristics. In the future we want to explore other prototypical aspects of urban legends like, for example, linguistic style \citep{Pennebaker99,louismakes}. With regard to sentiment, besides the simple word polarities we used,  we will explore the emotions expressed in UL, FT and GN, using an approach similar to the one described in \citep{carlo:rada:ACM-SAC-08}. 
Exploiting knowledge-based and corpus-based methods, that approach deals with automatic recognition of affect, annotating texts with six basic emotions.
We believe that fine-grained emotion annotation of urban legends could shed more light in the understanding the mechanisms behind persuasive language.

\bibliography{Persuasive}

\end{document}